%% file: main.tex
\documentclass[10pt,twocolumn,letterpaper]{article}

\usepackage{wacv}
\usepackage{times}
\usepackage{epsfig}
\usepackage{graphicx}
\usepackage{amsmath}
\usepackage{amssymb}
\usepackage{booktabs}
\usepackage{multirow}
\usepackage{hhline}
\usepackage{xcolor}
\usepackage{tikz, pgfplots}
\DeclareFixedFont{\ttb}{T1}{txtt}{bx}{n}{12} % for bold
\DeclareFixedFont{\ttm}{T1}{txtt}{m}{n}{12}  % for normal

\usepackage{color}
\definecolor{deepblue}{rgb}{0,0,0.5}
\definecolor{deepred}{rgb}{0.6,0,0}
\definecolor{deepgreen}{rgb}{0,0.5,0}

\usepackage{listings}

\newcommand\pythonstyle{\lstset{
language=Python,
basicstyle=\ttm,
otherkeywords={self},             % Add keywords here
keywordstyle=\ttb\color{deepblue},
emph={SpaceToDepthJIT,__call__,AADownsamplingJIT,__init__,FastGlobalAvgPool2d},          % Custom highlighting
emphstyle=\ttb\color{deepred},    % Custom highlighting style
stringstyle=\color{deepgreen},
frame=single,                         % Any extra options here
framexrightmargin=270pt,
showstringspaces=false            % 
}}

\lstnewenvironment{python_l}[1][]
{
\pythonstyle
\lstset{#1}
}
{}

\newcommand\pythoninline[1]{{\pythonstyle\lstinline!#1!}}
\usepackage{pythonhighlight}

\usepackage[page]{appendix}

\def\wacvPaperID{****} % Enter the WACV Paper ID here

\wacvfinalcopy % *** Uncomment this line for the final submission

\ifwacvfinal
\def\assignedStartPage{1} % *** Enter the assigned starting page number (instead of 9876)
\fi

\ifwacvfinal
\usepackage[breaklinks=true,bookmarks=false]{hyperref}
\else
\usepackage[pagebackref=true,breaklinks=true,colorlinks,bookmarks=false]{hyperref}
\fi

\ifwacvfinal
\else
\fi

\begin{document}

%%%%%%%%% TITLE
\title{TResNet: High Performance GPU-Dedicated Architecture}

\author{Tal Ridnik \hspace{0.1cm} Hussam Lawen  \hspace{0.1cm}  Asaf Noy  \hspace{0.1cm} Emanuel Ben Baruch \hspace{0.1cm} Gilad Sharir
\hspace{0.1cm} Itamar Friedman
\vspace{0.5cm} \\ 
DAMO Academy, Alibaba Group\\
{\tt\small $\{$tal.ridnik, hussam.lawen, asaf.noy,  emanuel.benbaruch, gilad.sharir, itamar.friedman$\}$}\\  {\tt\small @alibaba-inc.com }
}
\maketitle

\begin{abstract}
\input{abstract.tex}

\end{abstract}

\section{Introduction}
\input{introduction.tex}

\medskip
\section{TResNet Design}
\input{TResNet_Model.tex}

\section{ImageNet Results}
In this section, we will evaluate TResNet models on standard ImageNet training (input resolution 224), and compare their top-1 accuracy and GPU throughput to other known models. We will also perform an ablation study to better understand the effect of different refinements, show results for fine-tuning TResNet to higher input resolution, and do a thorough comparison to EfficientNet models.
\input{Results.tex}

\section{Transfer Learning Results}
In this section, we will present transfer learning results of TResNet models on four well-known single-label classification downstream datasets. We will also present transfer learning results on multi-label classification and object detection tasks.

\medskip
\subsection{Single-Label Classification}

\input{transfer_learning.tex}

\subsection{Multi-Label Classification}
\input{multi_label.tex}

\subsection{Object Detection}
\input{object_detection.tex}

\section{Conclusion}
\input{Conclusion.tex}

{\small
\bibliographystyle{ieee_fullname}
\bibliography{egbib}
}

\clearpage

\input{appendix}

\end{document}

%% file: abstract.tex
\label{sec:abstract}
Many deep learning models,  developed  in  recent  years, reach  higher ImageNet accuracy than ResNet50, with fewer or comparable FLOPs count. While FLOPs are often seen as a proxy for network efficiency, when measuring actual GPU training and inference throughput, vanilla ResNet50 is usually significantly faster than its recent competitors, offering better throughput-accuracy trade-off.

In this work, we introduce a series of architecture modifications that aim to boost neural networks' accuracy, while retaining their GPU training and inference efficiency. We first demonstrate and discuss the bottlenecks induced by FLOPs oriented optimizations. We then suggest alternative designs that better utilize GPU structure and assets. Finally, we introduce a new family of GPU-dedicated models, called TResNet, which achieves better accuracy and efficiency than previous ConvNets

Using a TResNet model, with similar GPU throughput to ResNet50, we reach $80.8\%$ top-1 accuracy on ImageNet. Our TResNet models also transfer well and achieve state-of-the-art accuracy on competitive single-label classification datasets such as Stanford Cars (96.0\%), CIFAR-10 (99.0\%), CIFAR-100 (91.5\%) and Oxford-Flowers (99.1\%).  TResNet models also achieve state-of-the-art results on a multi-label classification task, and  perform well on object detection.
Implementation is available at: https://github.com/mrT23/TResNet.

%% file: introduction.tex
\label{introduction} 

The seminal ResNet models \cite{he2016deep}, introduced in 2015, revolutionized the world of deep learning. ResNet models use repeated well-designed residual blocks, allowing training of very deep networks to high accuracy while maintaining high GPU utilization. ResNet models are also easy to train, and converge fast and consistent even with plain SGD optimizer \cite{yamazaki2019yet}. NVIDIA Volta tensor cores \cite{markidis2018nvidia} further improved ResNet models GPU utilization, up to quadrupling their GPU throughput on mixed-precision training and inference \cite{xu2018deep}. Among the ResNet models, ResNet50 established itself as a prominent model in terms of speed-accuracy trade-off, and became a leading backbone model for many computer vision tasks \cite{gao2019res2net,li2018detnet,xiao2018simple, johnson2018adapting}.

Since ResNet50, many modern deep learning models were developed, which achieve better ImageNet accuracy with fewer or comparable FLOPs.
Surprisingly, even though most deep learning models are trained, and sometimes deployed, on GPUs, few models try explicitly to find an optimal design in terms of GPU throughput. Since FLOPs are not an accurate proxy for GPU speed \cite{cai2018proxylessnas}, sub-optimal design for GPUs might occur. This is especially true for GPU training speed, which is rarely measured and documented in academic literature, and can be severely hindered by some modern architecture design tricks \cite{ma2018shufflenet}.
 
Table \ref{Table:models_comparison} compares ResNet50 to popular newer architectures, with similar ImageNet top-1  accuracy - ResNet50-D \cite{he2019bag}, ResNeXt50 \cite{ResNext}, SEResNeXt50 \cite{hu2018squeeze}, EfficientNet-B1 \cite{tan2019efficientnet} and MixNet-L \cite{tan2019mixnet}. 
\input{Tables/Models_comparison.tex}
We see from Table \ref{Table:models_comparison} that the reduction of FLOPs and the usage of new tricks in modern networks,
compared to ResNet50, is not translated to improvement in GPU throughput. This is especially evident for GPU training speed, where ResNet50 gives by a large margin better speed-accuracy trade-off.
We identify two main reasons for this throughput gap: \\
1.	Modern networks like EfficientNet, ResNeXt and MixNet do extensive usage of depthwise and 1x1 convolutions, that provide significantly fewer FLOPs than 3x3 convolutions. However, GPUs are usually limited by memory access cost and not by number of computations, especially for low-FLOPs layers. Hence, the reduction in FLOPs is not translated well to an equivalent increase in GPU throughput \cite{ma2018shufflenet}. \\
2. Modern networks like ResNeXt and MixNet do extensive usage of multi-path. For training, this creates lots of activation maps that need to be stored for backward propagation, which reduces the maximal possible batch size, thus hurting the GPU throughput. Multi-path also limits the ability to use inplace operations~\cite{inplaceABN}, and can lead to network fragmentation~\cite{ma2018shufflenet}.

Following our analysis of Table \ref{Table:models_comparison}, we want to design a new family of networks, TResNet, aimed at high accuracy while maintaining high GPU utilization. TResNet models will contain the latest published design tricks available, along with our own novelties and optimizations. Unlike previous works, which measure only the FLOPS proxy or just GPU inference speed, we will directly focus on both GPU inference and training speed.
For a proper comparison to previous models, one network variant (TResNet-M) is designed to match ResNet50 GPU throughput, while the rest match modern larger architectures. 

We will show that for all tested datasets, TResNet models offer an improved speed-accuracy trade-off. Specifically, they reach ImageNet top-1 accuracy of $80.8\%$ with GPU throughput similar to ResNet50 ($79.0\%$), and top-1 accuracy of $84.3\%$ with better GPU throughput than EfficientNet-B5 ($83.7\%$). Besides ImageNet, TResNet models also achieve state-of-the-art accuracy on 3 out of 4 widely used downstream single-label datasets, with x8-15 faster GPU inference speed. They also excel on multi-label classification and object detection tasks.

%% file: Tables/Models_comparison.tex
\begin{table*}
\centering
\begin{tabular}{|l|c|c|c|c|} 
\hline
Model & Training Speed [img/sec] & Inference Speed [img/sec] &  Top1 Accuracy [\%] &  Flops [G]  \\ 
\hline
ResNet50 \cite{he2016deep}                    & \textbf{805}                                                                      & 2830                                                                      & 79.0                                                        & 4.1                                                    \\ 
\hline
ResNet50-D \cite{he2019bag}                  & 600                                                                      & 2670                                                                      & 79.3                                                        & 4.4                                                    \\
ResNeXt50 \cite{ResNext}                   & 490                                                                      & 1940                                                                      & 79.4                                                        & 4.3                                                    \\
EfficientNetB1 \cite{tan2019efficientnet}              & 480                                                                      & 2740                                                                      & 79.2                                                        & 0.6                                                    \\
SEResNeXt50 \cite{tan2019efficientnet}                 & 400                                                                      & 1770                                                                      & 79.9                                                        & 4.3                                                    \\
MixNet-L \cite{tan2019mixnet}                    & 400                                                                      & 1400                                                                      & 79.0                                                        & 0.5                                                    \\ 
\hline
TResNet-M                   & 730                                                                      & \textbf{2930}                                                                      & \textbf{80.8}                                               & 5.5                                                    \\
\hline
\end{tabular}
\medskip
\caption{\textbf{Comparison of ResNet50 to top modern networks, with similar top-1 ImageNet accuracy}. All measurements were done on Nvidia V100 GPU with mixed precision. For gaining optimal speeds, training and inference were measured on 90\% of maximal possible batch size. Except TResNet-M, all the models' ImageNet scores were taken from the public repository \cite{RWightman}, which specialized in providing top implementations for modern networks. Except EfficientNet-B1, which has input resolution of 240, all other models have input resolution of 224.}
\label{Table:models_comparison}
\end{table*}

%% file: TResNet_Model.tex
\label{sec:TResnet_model}
TResNet design is based on the ResNet50 architecture, with dedicated refinements, modifications and optimizations. It contains three variants, TResNet-M, TResNet-L and TResNet-XL, that vary only in their depth and the number of channels. TResNet architecture contains the following refinements and changes compared to plain ResNet50 design: \begin{itemize}
 \setlength\itemsep{0.2pt}
\item SpaceToDepth Stem
\item Anti-Alias Downsampling
\item
In-Place Activated BatchNorm
\item Novel Block-type Selection 
\item Optimized SE Layers.
\end{itemize}
Previous works usually offer refinements to ResNet50 which increase the accuracy at the cost of reducing the GPU throughput \cite{he2019bag,lee2020compounding,hu2018squeeze,zhang2019shiftinvar}. Differently, in our design some refinements increase the models'
throughput and some decrease it. All-in-all, for TResNet-M we chose a mixture of refinements that provide a similar GPU throughput to ResNet50, for a fair comparison of the models' accuracy.  

\medskip
\subsection{Refinements}
\textbf{SpaceToDepth Stem} - Neural networks usually start with a stem unit - a component whose goal is to quickly reduce the input resolution. ResNet50 stem is comprised of a stride-2 conv7x7 followed by a max pooling layer \cite{he2016deep}, which reduces the input resolution by a factor of $4$ ($224 \rightarrow 56$). ResNet50-D stem design \cite{he2019bag}, for comparison, is more elaborate - the conv7x7 is replaced by three conv3x3 layers. The new ResNet50-D stem design did improve accuracy, but at a cost of lowering the training throughput - see Table \ref{Table:models_comparison}, where the new stem design is responsible for almost all the decline in the throughput.

\begin{figure} [hbt!]
  \centering
  \includegraphics[scale=.86]{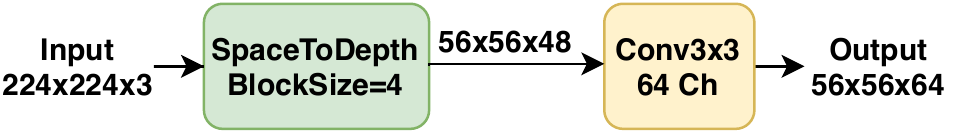}
  \caption{\textbf{TResNet-M SpaceToDepth stem design.}}
  \label{fig:sted_design_pic}
\end{figure}

We wanted to replace the traditional convolution-based downscaling unit by a fast and seamless layer, with little information loss as possible, and let the well-designed residual blocks do all the actual processing work. The new stem layer sole functionality should be to downscale the input resolution to match the rest of the architecture, e.g., by a factor of $4$. We met these goals by using a dedicated SpaceToDepth transformation layer \cite{space2depth_sandler2019non}, that rearranges blocks of spatial data into depth. Notice that in contrast to \cite{space2depth_sandler2019non}, which mainly used SpaceToDepth in the context of isometric (single-resolution) networks, in our novel design SpaceToDepth is used as a drop-in replacement for the tradition stem unit. The SpaceToDepth layer is followed by simple convolution, to match the number of wanted channels, as can be seen in Figure \ref{fig:sted_design_pic}.
\medskip

\textbf{Anti-Alias Downsampling (AA)} - \cite{zhang2019shiftinvar}  proposed to replace all downscaling layers in a network by an equivalent AA component, to improve the shift-equivariance of deep networks and give better accuracy and robustness. 

\begin{figure}[hbt!]
  \centering
  \includegraphics[scale=.35]{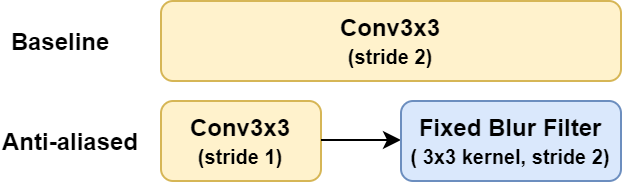}
  \caption{\textbf{The AA downsampling scheme of TResNet architecture}. All stride-2 convolutions are replaced by stride-1 convolutions, followed by a fixed downsampling blur filter \cite{zhang2019shiftinvar}.}
  \label{fig:anti-alias}
\end{figure}

We implemented an economic variant of AA, similar to \cite{lee2020compounding}, that provides an improved speed-accuracy trade-off - only our stride-2 convolutions are replaced by stride-1 convolutions followed by a 3x3 blur
kernel filter with stride 2, as described in Figure \ref{fig:anti-alias}.

\medskip
% \medskip

\textbf{In-Place Activated BatchNorm (Inplace-ABN)} - Along the architecture, we replaced all BatchNorm+ReLU layers by Inplace-ABN \cite{inplaceABN} layers, which implements BatchNorm with activation as a single inplace operation, allowing to reduce significantly the memory required for training deep networks, with only a small increase in the computational cost. As an activation function for the Inplace-ABN, we chose to use Leaky-ReLU instead of ResNet50's plain ReLU. 

Using Inplace-ABN in TResNet models offers the following advantages:
\begin{itemize}
\setlength\itemsep{0.1pt}
  \item BatchNorm layers are major consumers of GPU memory. Replacing BatchNorm layers with Inplace-ABN enables to significantly increase the maximal possible batch size, which can improve the GPU utilization.
  \item For TResNet models, Leaky-ReLU provides better accuracy than plain ReLU. While some modern activation, like Swish and Mish \cite{misra2019mish}, might also give better accuracy than ReLU, their GPU memory consumption is higher, as well as their computational cost. In contrast, Leaky-ReLU has exactly the same GPU memory consumption and computational cost as plain ReLU. 
  \item The increased batch size can also improve the effectiveness of popular algorithms like triplet loss \cite{Lawen_2020} and momentum-contrastive learning. \cite{he2019momentum} \\
\end{itemize}

\textbf{Novel Block-Type Selection} - ResNet34 and ResNet50 share the same architecture, with one difference: ResNet34 uses solely 'BasicBlock' layers, which comprise of two conv3x3 as the basic building block, while ResNet50 uses 'Bottleneck' layers, which comprise of two conv1x1 and one conv3x3 as the basic building block \cite{he2016deep}.  Bottleneck layers have higher GPU usage than BasicBlock layers, but usually give better accuracy. However, BasicBlock layers have larger receptive field, so they might be more suited to be placed at the early stages of a network.

We found that the uniform block selection of ResNet models is not optimal, and a better speed-accuracy trade-off can be obtained  using a novel design, which uses a mixture of BasicBlock and Bottleneck layers. Since BasicBlock layers have a larger receptive field, we placed them at the first two stages of the network, and Bottleneck layers at the last two stages.

Compared to ResNet50, we also modified the number of channels and the number of residual blocks in the 3rd stage for the different TResNet models. Full specification of TResNet networks, including block type, width and number of residual blocks in each stage, appears in Table \ref{Table:TResnet_layers_table}.

\input{Tables/TResnet_layers_table.tex}
% \medskip

\textbf{Optimized SE Layers} - We added dedicated squeeze-and-excitation~\cite{hu2018squeeze}  layers (SE) to TResNet architecture. In order to reduce the computational cost of the SE blocks, and gain the maximal speed-accuracy benefit, we placed SE layers only in the first three stages of the network. The last stage, which works on low-resolution maps, does not get a large accuracy benefit from the global average pooling operation that SE provides. 

\begin{figure}[hbt!]
  \centering
  \includegraphics[scale=.59]{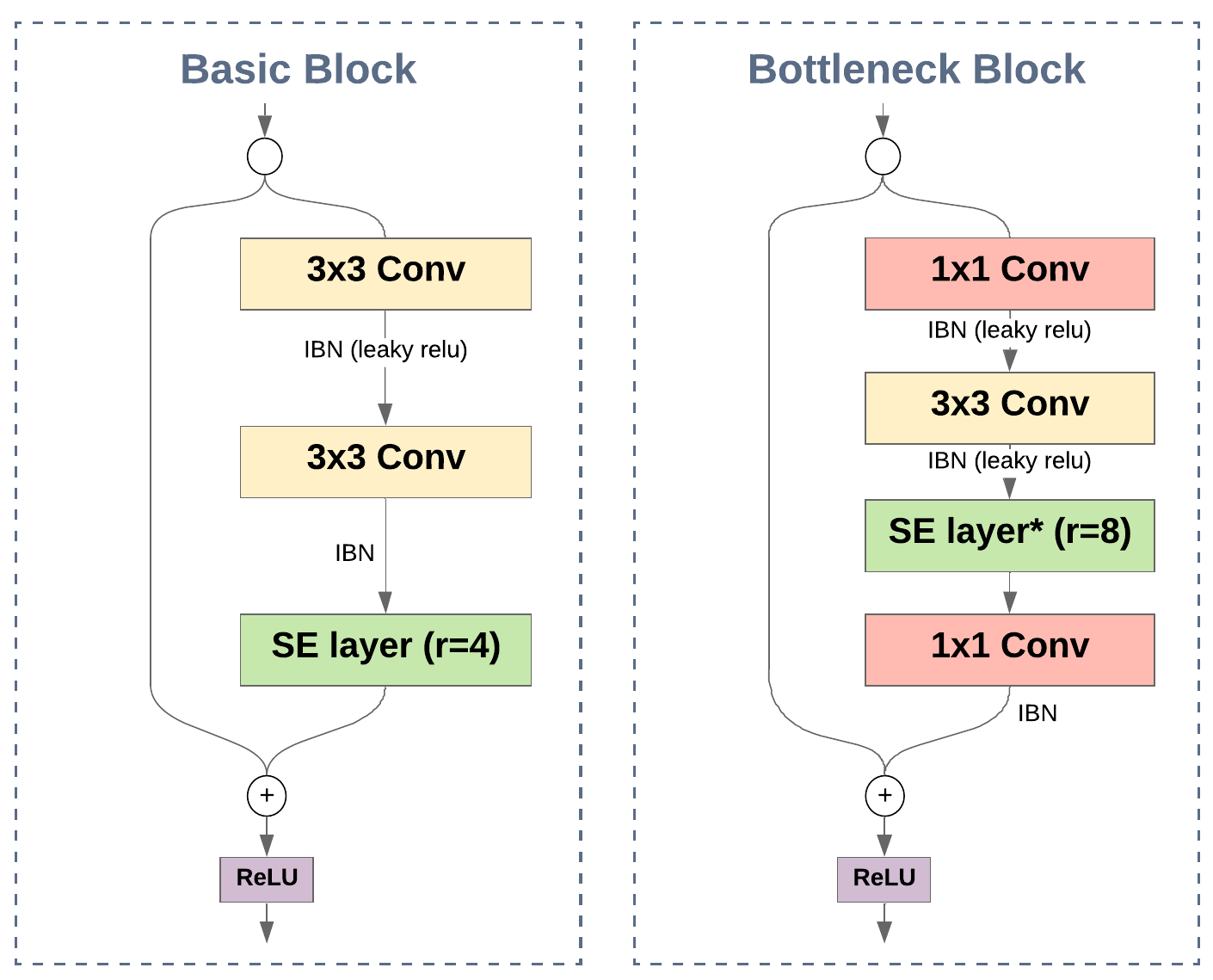}
  \caption{\textbf{TResNet BasicBlock and Bottleneck design} (stride 1). IBN = Inplace-BatchNorm, r = reduction factor, * - Only for 3rd stage.}
  \label{fig:bottleneck_block}
\end{figure}

Compared to standard SE design \cite{hu2018squeeze}, TResNet SE placement and hyper-parameters are also optimized: For Bottleneck units, we added the SE module after the conv3x3 operation, with a reduction factor of $8$, and for BasicBlock units, we added SE module just before the residual sum, with a reduction factor of $4$. This change aims to reduce the number of parameters and computational cost of SE layers: since BasicBlock units are placed on the first two stages of the network, they contain relatively low number of input channels, so only a small reduction factor ($4$) is needed. The Bottleneck units are placed in later stages of the network, with more input channels, so a higher reduction factor ($8$) is needed. Placing the SE layers after the reduction phase of the Bottleneck layer further reduces the computational cost.
The complete blocks design, with SE layers and Inplace-ABN, is presented in Figure \ref{fig:bottleneck_block}.
% \medskip
% \medskip

\subsection{Code Optimizations}
In this section we will describe code optimizations we did to enhance the GPU throughput and reduce the memory footprint of TResNet models. While code optimizations are sometimes overlooked and seen as 'implementation details', we claim that they are crucial for designing a modern network with top GPU performance.

We designed TResNet using the PyTorch library  \cite{paszke2017automatic}, due to its popularity and ability for easy code prototyping. However, all the optimization described below are also applicable to other deep learning libraries, such as TensorFlow \cite{abadi2016tensorflow} and MXNet \cite{chen2015mxnet} 

\subsubsection{JIT Compilation}
JIT compilation enables, at execution time, to dynamically compile a high-level code into highly-efficient, optimized 
machine code. This is in contrast to the default Pythonic option of running  a code dynamically, via an interpreter.
We used JIT compilations for network modules that don't contain learnable parameters - the AA blur filter and the SpaceToDepth modules. For modules without learnable parameters, JIT compilation is a seamless process that accelerates the network GPU throughput without imposing limitations on the actual training and inference - for example, the input size does not need to be fixed and pre-determined, flow control statements are still possible. 

For the AA and SpaceToDepth modules, we found that JIT compilation reduces the GPU cost by almost a factor of two. The module's JIT code appears in appendix \ref{sec:Modouls_code}.

\subsubsection{Inplace Operations}
Inplace operations change directly the content of a given tensor, without making a copy. They reduce the memory access cost of an operation, and also prevent creation of unneeded activation maps for backward propagation, hence increasing the maximal possible batch size.
In TResNet code, inplace operations are used as as much as possible. In addition of using Inplace-ABN, there are also inplace operations for the residual connection, SE layers, blocks' final activation and more. This is a key factor in enabling large batch size - TResNet-M maximal batch size is almost twice of ResNet50 - $512$, as can be seen in Table \ref{Table:models_comparison}. For full review of TResNet inplace operations, see our public code.

\subsubsection{Fast Global Average Pooling}
Global average pooling (GAP) is used heavily in TResNet architecture - both in the SE layers, and before the final fully connected.
GAP can be implemented via a general boilerplate method, AvgPool2d, that does average pooling of a tensor to a given spatial dimension, which in the GAP is (1,1). However, usually using this boilerplate methodology is usually not optimal, and leading to sub-optimal performances, mainly due to bad memory utilization.

We found that a simple dedicated implementation of GAP, with optimized code for the specific case of (1,1) spatial output, can be up to $5$ times faster than the boilerplate implementation on GPU, since it can bring data from memory more efficiently.
Our TResNet implementation for Fast GAP appears in the appendix \ref{sec:Modouls_code}.

\subsubsection{Deployment Optimizations}
There are dedicated optimizations for enhancing (frozen) model inference speed during deployment. For example, BatchNorm layers can be fully absorbed into the convolution layers before them, significantly accelerating the model. In addition, there are dedicated libraries for GPU deployment, such as TensorRT \cite{vanholder2016efficient}. However, we wanted to provide a fair comparison of TResNet to other architectures, while focusing on all aspects of GPU efficiency (training speed, inference speed and maximal batch size), so we avoided doing inference-tailored optimizations.
In practice, the inference speed reported for TResNet can be improved via such optimizations.

%% file: Tables/TResnet_layers_table.tex
\begin{table*}[hbt!]

\begin{center}

\begin{tabular}{|l|c|c|c|c|c|c|c|c|c|} 
\hline
{Layer} & {Block Type}                & {Output} & {Stride} & \multicolumn{6}{c|}{TResNet}\\ 
\cline{5-10}
                       & \multicolumn{1}{l|}{}                                           & \multicolumn{1}{l|}{}                        & \multicolumn{1}{l|}{}                                                    & \multicolumn{2}{c|}{M}                                                                            & \multicolumn{2}{c|}{L}                                                                            & \multicolumn{2}{c|}{XL}                                                                            \\ 
\cline{5-10}
                       & \multicolumn{1}{l|}{}                                           & \multicolumn{1}{l|}{}                        & \multicolumn{1}{l|}{}                                                    & Repeats                                           & Channels                                              & Repeats                                           & Channels                                              & Repeats                                           & Channels                                               \\ 
\hline
Stem                   & \begin{tabular}[c]{@{}c@{}}SpaceToDepth\\ Conv1x1 \end{tabular} & 56$\times$56                                 & \begin{tabular}[c]{@{}c@{}}-\\ 1 \end{tabular}                           & \begin{tabular}[c]{@{}c@{}}1\\ 1 \end{tabular} & \begin{tabular}[c]{@{}c@{}}48\\ 64 \end{tabular} & \begin{tabular}[c]{@{}c@{}}1\\ 1 \end{tabular} & \begin{tabular}[c]{@{}c@{}}48\\ 76 \end{tabular} & \begin{tabular}[c]{@{}c@{}}1\\ 1 \end{tabular} & \begin{tabular}[c]{@{}c@{}}48\\ 84 \end{tabular}  \\ 
\hline
Stage1                 & BasicBlock+SE                                                   & 56$\times$56                                 & 1                                                                        & 3                                              & 64                                               & 4                                              & 76                                               & 4                                              & 84                                                \\ 
\hline
Stage2                 & BasicBlock+SE                                                   & 28$\times$28                                 & 2                                                                        & 4                                              & 128                                              & 5                                              & 152                                              & 5                                              & 168                                               \\ 
\hline
Stage3                 & Bottleneck+SE                                                   & 14$\times$14                                 & 2                                                                        & 11                                             & 1024                                             & 18                                             & 1216                                             & 24                                             & 1344                                              \\ 
\hline
Stage4                 & Bottleneck                                                      & 7$\times$7                                 & 2                                                                        & 3                                              & 2048                                             & 3                                              & 2432                                             & 3                                              & 2688                                              \\ 
\hline
Pooling                & GlobalAvgPool                                                   & 1$\times$1                                   & 1                                                                        & 1                                              & 2048                                             & 1                                              & 2432                                            & 1                                              & 2688                                             \\ 
\hline\hline

\#Params.          & \multicolumn{3}{c|}{}                                                                                                                                                                     & \multicolumn{2}{c|}{29.4M}                                                                        & \multicolumn{2}{c|}{54.7M}                                                                       & \multicolumn{2}{c|}{77.1M}                                                                        \\
\hline
\end{tabular}
\end{center}
% \medskip
\caption{\textbf{Overall architecture of the three TResNet models.}}
\label{Table:TResnet_layers_table}
\end{table*}

%% file: Results.tex
\medskip
\subsection{Basic Training}
Our main benchmark for evaluating TResNet models is the popular ImageNet dataset \cite{NIPS2012_4824}. 
We trained the models on input resolution  $224$, for $300$
epochs, using a SGD optimizer and 1-cycle policy \cite{smith2018disciplined}. For regularization, we used AutoAugment \cite{cubuk2019autoaugment}, Cutout \cite{Cutout}, Label-smoothing \cite{DBLP:journals/corr/SzegedyVISW15} and True-weight-decay \cite{loshchilov2017decoupled}. We found that the common ImageNet statistics normalization \cite{lee2020compounding,cubuk2019autoaugment,tan2019efficientnet} does not improve the training accuracy, and instead normalized all the RGB channels to be between $0$ and $1$. For comparison, we repeated the same training procedure for ResNet50. Results appear in Table \ref{Table:TResNet scores}. 

\medskip
\input{Tables/tresnet_imagenet_results} \medskip

We can see from Table \ref{Table:TResNet scores} that TResNet-M, which has similar GPU throughput to ResNet50, has significantly higher validation accuracy on ImageNet ($+1.8\%$). It also outperforms all the other models that appear in Table \ref{Table:models_comparison}, both in terms of GPU throughput and ImageNet top-1 accuracy. 

Note that our ResNet50 ImageNet top-1 accuracy, $79.0\%$, is significantly higher than the accuracy stated in previous articles \cite{he2016deep,he2019bag, kolesnikov2019large}, demonstrating the effectiveness of our training procedure. In addition, training TResNet-M and ResNet50 models takes less than 24 hours on an 8xV100 GPU machine, showing that our training scheme is also efficient and economical.

Another strength of the TResNet models, as reflected by Table \ref{Table:TResNet scores}, is the ability to work with significantly larger batch sizes than other models. In general, large batch size leads to better GPU utilization, and allows easier scaling to large inputs. For distributed learning, it also reduces the number of synchronization needed in an epoch between the different GPUs.

\medskip
\subsection{Ablation Study}
\subsubsection{Network Refinements}

We performed an ablation study to evaluate the impact of the different refinements and modifications in TResNet-M model on the validation accuracy, inference speed, training speed and maximal batch size. Results appear in Table \ref{Table:ablation_study}.
We can better understand from Table \ref{Table:ablation_study} the contribution of each refinement:

\input{Tables/ablation}

\textbf{SpaceToDepth}: The SpaceToDepth module provides improvements to all the indices. Notice that while we are not the first to use this innovative module (see \cite{space2depth_sandler2019non}), we are the first to  integrate it into a high-performance network as drop-in replacement for the traditional convolution-based stem, and get a meaningful improvement.
While the GPU throughput improvement is expected, the fact that also the accuracy improves (marginally) when replacing the ResNet stem cell by a "cheaper" SpaceToDepth unit is somewhat surprising. This result supports our intuition that there could be "information loss" within convolution-based stem unit - details from the original image are not propagated well due to the aggressive downscaling process. Although simpler, a SpaceToDepth module minimizes this loss, and enables to process the data via the residual blocks, which are protected from information loss by the skip connections. We will further investigate this issue in future works.

\textbf{Block-Type selection}:
Our novel block-type selection scheme, which uses both 'BasicBlock' and 'Bottleneck' blocks in the network, provides significant improvements to all indices, and shows that the uniform block-type design of the ResNet models is not optimal. 
Notice that this refinement also includes changing the number of residual blocks in the 3rd stage, from $6$ to $11$. In practice, the actual number ($11$) was chosen after all the other refinements were finalized. Its goal was to bring TResNet-M to a similar GPU throughput as ResNet50. Setting the number of blocks in the 3rd stage to $11$ gave a slightly lower training speed and a slightly higher inference speed than ResNet50. However, using fewer blocks at the 3rd stage could have given relative improvement also for the training speed, at the cost of reducing the accuracy.

\textbf{Inplace-ABN}:
As expected, Inplace-ABN enables us to significantly increase the possible batch size, by $200$ images. In addition, using Leaky-ReLU as activation function, instead of plain ReLU, marginally improved the accuracy. However, in terms of contribution to the actual GPU throughput, the impact of Inplace-ABN is mixed: while the inference speed improved, the training speed was somewhat reduced. While Inplace-ABN enables us to increase the batch size (which should be translated to better GPU throughput), it is also a more complicated module than a simple BatchNorm layer, so there is an inherent trade-off for using it. Since the ability to work with larger batch is highly beneficial for multi-GPU training and some dedicated loss functions,  we chose to include this refinement.

\textbf{Optimized SE  + Anti-Aliasing layers}: As expected, these layers significantly improve the ImageNet top-1 accuracy,  with a price of reducing the model GPU throughput. We were able to compensate for this decrease with the previous refinements, and all-in-all get a better speed-accuracy trade-off than ResNet50.

\subsubsection{Code Optimizations}
In Table \ref{Table:code_ablation_study} we evaluate the contribution of the different code optimizations.
\input{Tables/code_ablation}
We can see from Table \ref{Table:code_ablation_study} that among the optimizations, dedicated inplace operations give the greatest boost - not only it improves the GPU throughput, but it also significantly increases the maximal possible batch size, since it avoids the creation of unneeded activation maps for backward propagation.

\subsection{High-Resolution Fine-Tuning}
We tested the scaling of TResNet models to higher input resolutions on ImageNet. We used the pre-trained TResNet models that appear in Table \ref{Table:TResNet scores} as a starting point, and did a short $10$ epochs fine-tuning to input resolution of $448$. The results appear in Table \ref{Table:high_resolution_imagenet}.

\input{Tables/high_resolution_imagenet}

We see from Table \ref{Table:high_resolution_imagenet} that TResNet models scale well to high resolutions. Even TResNet-M, which is a relatively small and compact model, can achieve top-1 accuracy of $83.2\%$ on ImageNet with high-resolution input.  TResNet largest variant, TResNet-XL, achieves $84.3\%$ top-1 accuracy on ImageNet. 

\medskip
\subsection{Comparison to EfficientNet Models}
EfficientNet models, which are based on MobilenetV3 architecture \cite{howard2019searching}, propose to balance the resolution, height, and width of a base network for generating a series of larger networks. They are considered state-of-the-art architectures, that provide efficient networks for all ImageNet top-1 accuracy spectrum \cite{tan2019efficientnet}.
In Figure \ref{Table:EfficienetNetVsTResNetInference} and Figure \ref{Table:EfficienetNetVsTResNetTraining}, we compare the inference and training speed of TResNet models to the different EfficientNet models respectively.
\input{Tables/EfficienetNetVsTResNetInference} 
\input{Tables/EfficienetNetVsTResNetTraining}

We can see from Figure \ref{Table:EfficienetNetVsTResNetInference} and Figure \ref{Table:EfficienetNetVsTResNetTraining} that all along the top-1 accuracy curve, TResNet models give better inference-speed-accuracy and training-speed-accuracy tradeoff than EfficientNet models. Note that each EfficientNet model was bundled and optimized to a specific resolution, while TResNet models were trained and tested on multi-resolutions, which makes this comparison biased toward EfficientNet models; Yet, TResNet models show superior results. Also note that EfficientNet models were trained for $450$ epochs and not for $300$ epochs like TResNet models, and that EfficientNet training procedure included more GPU intensive tricks (RMSProp  optimizer, drop-block) \cite{tan2019efficientnet}, so the actual gap in training times is even higher than stated in Figure \ref{Table:EfficienetNetVsTResNetTraining}.

%% file: Tables/tresnet_imagenet_results.tex
\begin{table}[hbt!]
\centering
\begin{tabular}{|l|c|c|c|c|} 
\hline
\multicolumn{1}{|c|}{Models} & \begin{tabular}[c]{@{}c@{}}Top\\Training\\Speed\\(img/sec)\end{tabular} & \begin{tabular}[c]{@{}c@{}}Top\\Inference\\Speed\\(img/sec)\end{tabular} & \begin{tabular}[c]{@{}c@{}}Max\\Train\\Batch\\Size\end{tabular} & \begin{tabular}[c]{@{}c@{}}Top-1\\Acc.\\{[}\%]\end{tabular}  \\ 
\hline
ResNet50                     & \textbf{805}                                                            & 2830                                                                     & 288                                                                    & 79.0                                                        \\ 
\hline
% EfficientNetB1               & 480                                                                     & 2740                                                                     & 196                                                                    & 79.2                                                        \\ 
% \hline
TResNet-M                    & 730                                                                     & \textbf{2930}                                                            & \textbf{512}                                                           & 80.8                                                        \\
TResNet-L                    & 345                                                                     & 1390                                                                     & 316                                                                    & 81.5                                                        \\
TResNet-XL                   & 250                                                                     & 1060                                                                     & 240                                                                    & \textbf{82.0}                                               \\
\hline
\end{tabular}
\medskip
\caption{\textbf{TResNet models accuracy and GPU throughput on ImageNet, compared to ResNet50}. All measurements were done on Nvidia V100 GPU, with mixed precision. All models are trained on input resolution of 224.}
\label{Table:TResNet scores}
% \vspace{-1mm}
\end{table}

%% file: Tables/ablation.tex
% \begin{table}[hbt!]
%     \begin{center}
%         \begin{tabular}{l|c|c}
%         \hline
%             \vtop{\hbox{Refinement}}  &          \multicolumn{1}{p{1.38cm}|}{\centering Top-1 Accuracy} &
%             \multicolumn{1}{p{1.34cm}}{\centering Inference speed (img/sec)} \\
           
%             \hline
%             Original ResNet50
%             & $79.0$ 
%             & 2830 \\
%             \hline
%             + Stem $\rightarrow$ SpaceToDepth
%             & $79.1$ 
%             & 2950 \\
%             \hline
%             + Block-type selection
%             & $79.4$ 
%             & 3320 \\
%             \hline
%             + Inplace-ABN
%             & $79.5$ 
%             & 3470 \\
%             \hline
%             + Optimizer SE layers
%             & $80.3$  
%             & 3280 \\
%             \hline
%             + AA
%             & $80.8$                
%             & 2930 \\
%           \hline
%         \end{tabular}
%     \end{center}
%     \caption{\textbf{Ablation study} - The impact of refinements in TResNet-M model on ImageNet top-1 accuracy and inference speed.}
%     \label{Table:ablation_study}
% \vspace{-3mm}    
% \end{table}

\begin{table*}[hbt!]
\centering
\begin{tabular}{c|c|c|c|c} 
\hline
Refinement         & Top1 Accuracy [\%]                                        & \begin{tabular}[c]{@{}c@{}}Inference Speed\\{[}img/sec]\end{tabular} & \begin{tabular}[c]{@{}c@{}}Training Speed\\{[}img/sec]\end{tabular} & \begin{tabular}[c]{@{}c@{}}Maximal \\{}Batch size\end{tabular}           \\ 
\specialrule{.1em}{.05em}{.05em}
ResNet50          & 79.0                                                 & 2830                                                                 & 805                                                                 & 288                                                  \\ 
\specialrule{.1em}{.05em}{.05em}
+ SpaceToDepth     & \begin{tabular}[c]{@{}c@{}}79.1\\(+0.1)\end{tabular} & \begin{tabular}[c]{@{}c@{}}2950\\(+120)\end{tabular}                 & \begin{tabular}[c]{@{}c@{}}830\\(+25)\end{tabular}                  & \begin{tabular}[c]{@{}c@{}}312\\(+24)\end{tabular}   \\ 
\hline
+ Novel Block-Type Selection  & \begin{tabular}[c]{@{}c@{}}79.4\\(+0.3)\end{tabular} & \begin{tabular}[c]{@{}c@{}}3320\\(+370)\end{tabular}                 & \begin{tabular}[c]{@{}c@{}}930\\(+100)\end{tabular}                 & \begin{tabular}[c]{@{}c@{}}424\\(+112)\end{tabular}  \\ 
\hline
+ Inplace-ABN      & \begin{tabular}[c]{@{}c@{}}79.5\\(+0.1)\end{tabular} & \begin{tabular}[c]{@{}c@{}}3470\\(+150)\end{tabular}                 & \begin{tabular}[c]{@{}c@{}}880\\(-50)\end{tabular}                  & \begin{tabular}[c]{@{}c@{}}624\\(+200)\end{tabular}  \\ 
\hline
+ Optimized SE     & \begin{tabular}[c]{@{}c@{}}80.3\\(+0.8)\end{tabular} & \begin{tabular}[c]{@{}c@{}}3280\\(-190)\end{tabular}                 & \begin{tabular}[c]{@{}c@{}}815\\(-65)\end{tabular}                  & \begin{tabular}[c]{@{}c@{}}596\\(-32)\end{tabular}   \\ 
\hline
+ AA               & \begin{tabular}[c]{@{}c@{}}80.8\\(+0.5)\end{tabular} & \begin{tabular}[c]{@{}c@{}}2930\\(-350)\end{tabular}                 & \begin{tabular}[c]{@{}c@{}}730\\(-85)\end{tabular}                  & \begin{tabular}[c]{@{}c@{}}512\\(-80)\end{tabular}   \\ 
% \specialrule{.1em}{.05em}{.05em}
% Total (TResNet-M) & 80.8                                                 & 2930                                                                 & 730                                                                 & 512                                                  \\
\hline
\end{tabular}
\vspace{+3mm} 
\caption{\textbf{Ablation study - The impact of different refinements in TResNet-M model on ImageNet top-1 accuracy, inference speed, training speed and maximal batch size}. All measurements are done on Nvidia V100 GPU, with mixed precision.}
\label{Table:ablation_study}

\end{table*}

%% file: Tables/code_ablation.tex
\begin{table}[hbt!]
\centering
\begin{tabular}{c|c|c|c} 
\hline
Refinement                                                                 & \begin{tabular}[c]{@{}c@{}}Inference \\Speed\\{[}img/sec] \end{tabular} & \begin{tabular}[c]{@{}c@{}}Training \\Speed\\{[}img/sec] \end{tabular} & \begin{tabular}[c]{@{}c@{}}Maximal \\Batch \\size \end{tabular}  \\ 
\specialrule{.1em}{.05em}{.05em}
\begin{tabular}[c]{@{}c@{}}TResNet-M\\No Code \\Optimizations\end{tabular} & 2790                                                                    & 670                                                                    & 420                                                              \\ 
\specialrule{.1em}{.05em}{.05em}
+ JIT                                                                      & \begin{tabular}[c]{@{}c@{}}2830\\(+40) \end{tabular}                    & \begin{tabular}[c]{@{}c@{}}695\\(+25) \end{tabular}                    & \begin{tabular}[c]{@{}c@{}}420\\(0) \end{tabular}                \\ 
\hline
\begin{tabular}[c]{@{}c@{}}+ Inplace\\Operations\end{tabular}              & \begin{tabular}[c]{@{}c@{}}2910\\(+80) \end{tabular}                    & \begin{tabular}[c]{@{}c@{}}715\\(+20) \end{tabular}                    & \begin{tabular}[c]{@{}c@{}}512\\(+92) \end{tabular}              \\ 
\hline
\begin{tabular}[c]{@{}c@{}}+ Fast\\ GAP\end{tabular}                & \begin{tabular}[c]{@{}c@{}}2930\\(+20) \end{tabular}                    & \begin{tabular}[c]{@{}c@{}}730\\(+15) \end{tabular}                     & \begin{tabular}[c]{@{}c@{}}512\\(0) \end{tabular}                \\
\hline
\end{tabular}
\medskip
\caption{\textbf{Ablation study - The impact of code optimizations in TResNet-M model on inference speed, training speed and maximal batch size}.}
\label{Table:code_ablation_study}
\end{table}

%% file: Tables/high_resolution_imagenet.tex
\begin{table}[hbt!]
\begin{center}
\begin{tabular}{|c|c|c|}
\hline
Model     & \begin{tabular}[c]{@{}c@{}}Input \\ Resolution\end{tabular} & \begin{tabular}[c]{@{}c@{}}Top-1 \\ Accuracy {[}\%{]}\end{tabular} 
\\ \hline
TResNet-M  & 224              & 80.8                                                                           \\ 
TResNet-M  & 448              & 83.2                                                                           \\ \hline
TResNet-L  & 224              & 81.5                                                                           \\ 
TResNet-L  & 448              & 83.8                                                                           \\ \hline
TResNet-XL & 224              & 82.0                                                                           \\ 
TResNet-XL & 448              & \textbf{84.3}                                                                           \\ \hline

% EfficieneNet-B5 & 456              & 83.7                                                                           \\ \hline
\end{tabular}
\end{center}
\caption{\textbf{Impact of the input resolution on the top1 ImageNet accuracy for TResNet models}. All TResNet $448$ input-resolution accuracies are obtained with $10$ epochs of fine-tuning. }
\label{Table:high_resolution_imagenet}
% \vspace{-1mm}
\end{table}

%% file: Tables/EfficienetNetVsTResNetInference.tex
\begin{figure}[hbt!]

\begin{tikzpicture}[scale = 1.08]
\pgfplotscreateplotcyclelist{mycolorlist}{%
blue,every mark/.append style={fill=blue!80!black},mark=*\\%
red,every mark/.append style={fill=red!80!black},mark=square*\\%
brown!60!black,every mark/.append style={fill=brown!80!black},mark=otimes*\\%
black,mark=star\\%
blue,every mark/.append style={fill=blue!80!black},mark=diamond*\\%
red,densely dashed,every mark/.append style={solid,fill=red!80!black},mark=*\\%
brown!60!black,densely dashed,every mark/.append style={
solid,fill=brown!80!black},mark=square*\\%
black,densely dashed,every mark/.append style={solid,fill=gray},mark=otimes*\\%
blue,densely dashed,mark=star,every mark/.append style=solid\\%
red,densely dashed,every mark/.append style={solid,fill=red!80!black},mark=diamond*\\%
}
\begin{axis}[
	xlabel={Inference Speed (images/sec)},
	title= {\bf{Top-1 Accuracy [\%] Vs  Inference Speed}},
	cycle list name=mycolorlist
]
% \addplot [mark=triangle*,
%     visualization depends on=\thisrow{alignment} \as \alignment,
%     nodes near coords, % Place nodes near each coordinate
%     point meta=explicit symbolic, % The meta data used in the nodes is not explicitly provided and not numeric
%     every node near coord/.style={anchor=\alignment} % Align each coordinate at the anchor 40 degrees clockwise from the right edge
%     ] table [% Provide data as a table
%      meta index=2 % the meta data is found in the third column
%      ] {
% x       y       label       alignment
% 2830    79.0    ResNet50      -360
% };

% \addplot [mark=*,
%     visualization depends on=\thisrow{alignment} \as \alignment,
%     nodes near coords, % Place nodes near each coordinate
%     point meta=explicit symbolic, % The meta data used in the nodes is not explicitly provided and not numeric
%     every node near coord/.style={anchor=\alignment} % Align each coordinate at the anchor 40 degrees clockwise from the right edge
%     ] table [% Provide data as a table
%      meta index=2 % the meta data is found in the third column
%      ] {
% x       y       label       alignment
% 2670    79.3    ResNet50-D      -150
% };
\addplot [mark=square*,  color=blue,
    visualization depends on=\thisrow{alignment} \as \alignment,
    nodes near coords, % Place nodes near each coordinate
    point meta=explicit symbolic, % The meta data used in the nodes is not explicitly provided and not numeric
    every node near coord/.style={anchor=\alignment} % Align each coordinate at the anchor 40 degrees clockwise from the right edge
    ] table [% Provide data as a table
     meta index=2 % the meta data is found in the third column
     ] {
x       y       label       alignment
2740    79.2    B1@240      0
1870    80.3    B2@260      0
1076    81.7    B3@300      0
565     83.0    B4@380      90
266     83.7    B5@456      90
};
% \addplot [mark=square*, color=blue,
%     visualization depends on=\thisrow{alignment} \as \alignment,
%     nodes near coords, % Place nodes near each coordinate
%     point meta=explicit symbolic, % The meta data used in the nodes is not explicitly provided and not numeric
%     every node near coord/.style={anchor=\alignment} % Align each coordinate at the anchor 40 degrees clockwise from the right edge
%     ] table [% Provide data as a table
%      meta index=2 % the meta data is found in the third column
%      ] {
% x       y       label       alignment
% 1870    80.3    EfficientNet-B2      -130
% };
% \addplot [mark=x,
%     visualization depends on=\thisrow{alignment} \as \alignment,
%     nodes near coords, % Place nodes near each coordinate
%     point meta=explicit symbolic, % The meta data used in the nodes is not explicitly provided and not numeric
%     every node near coord/.style={anchor=\alignment} % Align each coordinate at the anchor 40 degrees clockwise from the right edge
%     ] table [% Provide data as a table
%      meta index=2 % the meta data is found in the third column
%      ] {
% x       y       label       alignment
% 1770    79.0    SEResNeXt50      -160
% };
\addplot [mark=asterisk, color=red,
    visualization depends on=\thisrow{alignment} \as \alignment,
    nodes near coords, % Place nodes near each coordinate
    point meta=explicit symbolic, % The meta data used in the nodes is not explicitly provided and not numeric
    every node near coord/.style={anchor=\alignment} % Align each coordinate at the anchor 40 degrees clockwise from the right edge
    ] table [% Provide data as a table
     meta index=2 % the meta data is found in the third column
     ] {
x       y       label       alignment
2930    80.8    \textbf{M@224}      290
1398    81.5    \textbf{L@224}      200
1060    82.0    \textbf{XL@224}      200
790    83.2    \textbf{M@448}      200
360    83.8    \textbf{L@448}      200
258     84.3     \textbf{XL@448} 200
};
\legend{EfficientNet,TResNet}
\end{axis}

\end{tikzpicture}

\caption{\textbf{TResNet Vs EfficientNet models inference speed comparison.} Y label is the accuracy[\%]}
\label{Table:EfficienetNetVsTResNetInference}

\end{figure}
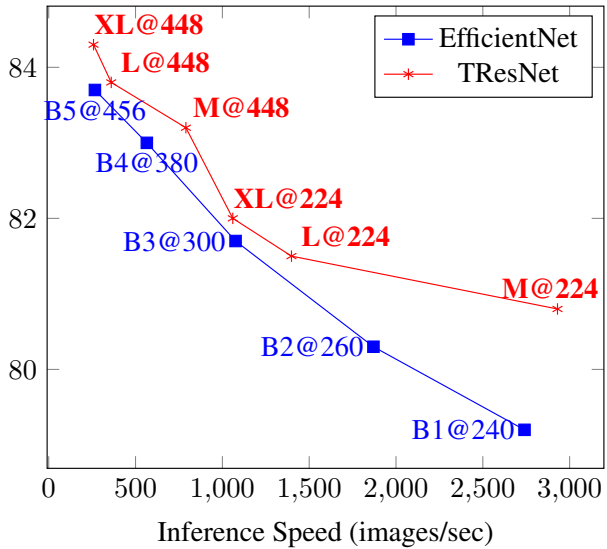

%% file: Tables/EfficienetNetVsTResNetTraining.tex
\begin{figure}[hbt!]
\begin{tikzpicture}[scale = 1.08]
\pgfplotscreateplotcyclelist{mycolorlist}{%
blue,every mark/.append style={fill=blue!80!black},mark=*\\%
red,every mark/.append style={fill=red!80!black},mark=square*\\%
brown!60!black,every mark/.append style={fill=brown!80!black},mark=otimes*\\%
black,mark=star\\%
blue,every mark/.append style={fill=blue!80!black},mark=diamond*\\%
red,densely dashed,every mark/.append style={solid,fill=red!80!black},mark=*\\%
brown!60!black,densely dashed,every mark/.append style={
solid,fill=brown!80!black},mark=square*\\%
black,densely dashed,every mark/.append style={solid,fill=gray},mark=otimes*\\%
blue,densely dashed,mark=star,every mark/.append style=solid\\%
red,densely dashed,every mark/.append style={solid,fill=red!80!black},mark=diamond*\\%
}
\begin{axis}[
	xlabel={Training Speed (images/sec)},
	title= {\bf{Top-1 Accuracy [\%] Vs  Training Speed}},
	cycle list name=mycolorlist
]
% \addplot [mark=triangle*,
%     visualization depends on=\thisrow{alignment} \as \alignment,
%     nodes near coords, % Place nodes near each coordinate
%     point meta=explicit symbolic, % The meta data used in the nodes is not explicitly provided and not numeric
%     every node near coord/.style={anchor=\alignment} % Align each coordinate at the anchor 40 degrees clockwise from the right edge
%     ] table [% Provide data as a table
%      meta index=2 % the meta data is found in the third column
%      ] {
% x       y       label       alignment
% 2830    79.0    ResNet50      -360
% };

% \addplot [mark=*,
%     visualization depends on=\thisrow{alignment} \as \alignment,
%     nodes near coords, % Place nodes near each coordinate
%     point meta=explicit symbolic, % The meta data used in the nodes is not explicitly provided and not numeric
%     every node near coord/.style={anchor=\alignment} % Align each coordinate at the anchor 40 degrees clockwise from the right edge
%     ] table [% Provide data as a table
%      meta index=2 % the meta data is found in the third column
%      ] {
% x       y       label       alignment
% 2670    79.3    ResNet50-D      -150
% };
\addplot [mark=square*,  color=blue,
    visualization depends on=\thisrow{alignment} \as \alignment,
    nodes near coords, % Place nodes near each coordinate
    point meta=explicit symbolic, % The meta data used in the nodes is not explicitly provided and not numeric
    every node near coord/.style={anchor=\alignment} % Align each coordinate at the anchor 40 degrees clockwise from the right edge
    ] table [% Provide data as a table
     meta index=2 % the meta data is found in the third column
     ] {
x       y       label       alignment
480    79.2    B1@240      0
400    80.3    B2@260      0
242    81.7    B3@300      0
110    83.0    B4@380      90
42     83.7    B5@456     90
};
% \addplot [mark=square*, color=blue,
%     visualization depends on=\thisrow{alignment} \as \alignment,
%     nodes near coords, % Place nodes near each coordinate
%     point meta=explicit symbolic, % The meta data used in the nodes is not explicitly provided and not numeric
%     every node near coord/.style={anchor=\alignment} % Align each coordinate at the anchor 40 degrees clockwise from the right edge
%     ] table [% Provide data as a table
%      meta index=2 % the meta data is found in the third column
%      ] {
% x       y       label       alignment
% 1870    80.3    EfficientNet-B2      -130
% };
% \addplot [mark=x,
%     visualization depends on=\thisrow{alignment} \as \alignment,
%     nodes near coords, % Place nodes near each coordinate
%     point meta=explicit symbolic, % The meta data used in the nodes is not explicitly provided and not numeric
%     every node near coord/.style={anchor=\alignment} % Align each coordinate at the anchor 40 degrees clockwise from the right edge
%     ] table [% Provide data as a table
%      meta index=2 % the meta data is found in the third column
%      ] {
% x       y       label       alignment
% 1770    79.0    SEResNeXt50      -160
% };
\addplot [mark=asterisk, color=red,
    visualization depends on=\thisrow{alignment} \as \alignment,
    nodes near coords, % Place nodes near each coordinate
    point meta=explicit symbolic, % The meta data used in the nodes is not explicitly provided and not numeric
    every node near coord/.style={anchor=\alignment} % Align each coordinate at the anchor 40 degrees clockwise from the right edge
    ] table [% Provide data as a table
     meta index=2 % the meta data is found in the third column
     ] {
x       y       label       alignment
730    80.8    \textbf{M@224}      290
345    81.5    \textbf{L@224}      200
250    82.0    \textbf{XL@224}      200
190    83.2    \textbf{M@448}      200
89     83.8    \textbf{L@448}      200
56     84.3     \textbf{XL@448} 200
};
\legend{EfficientNet,TResNet}
\end{axis}

\end{tikzpicture}
\caption{\textbf{TResNet Vs EfficientNet models training speed comparison.} Y label is the accuracy[\%]}
\label{Table:EfficienetNetVsTResNetTraining}

\end{figure}
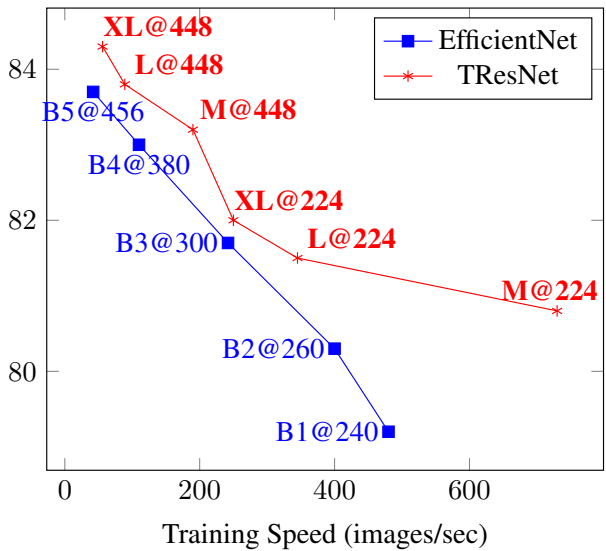

%% file: transfer_learning.tex
We evaluated TResNet on four commonly used, competitive transfer learning datasets: Stanford-Cars \cite{stanford_cars}, CIFAR-10 \cite{cifar}, CIFAR-100 \cite{cifar} and Oxford-Flowers \cite{oxford_flowers}.
For each dataset, we used ImageNet pre-trained checkpoints, and fine-tuned the models for 80 epochs using 1-cycle policy \cite{smith2018disciplined} . For the fine-grained classification tasks (Stanford-Cars and  Oxford-Flowers), in addition to cross-entropy loss we used weighted triplet loss with soft-margin \cite{ristani2018features,Lawen_2020}, which emphasizes hard examples by focusing of the most difficult positives and negatives samples in the batch. Table \ref{Table:transfer_table} shows the transfer learning performance of TResNet, compared to the known state-of-the-art models. 

\medskip

\input{Tables/transfer_table.tex}
\medskip

We can see from Table \ref{Table:transfer_table} that TResNet surpasses or matches the state-of-the-art accuracy on $3$ of the $4$ datasets, with x8-15 faster GPU inference speed. Note that all TResNet's results are from single-crop single-model evaluation.

%% file: Tables/transfer_table.tex
\begin{table}[hbt!]
\centering
\begin{tabular}{|l|l|c|c|c|} 
\hline
                            Dataset  & Model                                             &\begin{tabular}[c]{@{}c@{}}Top-1\\Acc.\end{tabular}          & \begin{tabular}[c]{@{}c@{}}Speed\\img/sec\end{tabular} & Input  \\ 
\hline
\multirow{2}{*}{\small CIFAR-10}      & {\small Gpipe}& \textbf{99.0} & -                                                      & 480    \\ 
\cline{2-5}
                              & {\small TResNet-XL}                                               & \textbf{99.0} & \textbf{1060}                                          & 224    \\ 
\hhline{|=====|}
\multirow{2}{*}{\small CIFAR-100}     & {\small EfficientNet-B7}                                          & \textbf{91.7} & 70                                                     & 600    \\ 
\cline{2-5}
                              & {\small TResNet-XL}                                               & 91.5          & \textbf{1060}                                          & 224    \\ 
\hhline{|=====|}
\multirow{2}{*}{\begin{tabular}[c]{@{}l@{}}\small Stanford \\\small Cars\end{tabular}} & {\small EfficientNet-B7}                                          & 94.7          & 70                                                     & 600    \\ 
\cline{2-5}
                              & {\small TResNet-L}                                                & \textbf{96.0} & \textbf{500}                                           & 368    \\ 
\hhline{|=====|}
\multirow{2}{*}{\begin{tabular}[c]{@{}l@{}}\small Oxford-\\\small Flowers\end{tabular}}       & {\small EfficientNet-B7}                                          & 98.8          & 70                                                     & 600    \\ 
\cline{2-5}
                              & {\small TResNet-L}                                                & \textbf{99.1} & \textbf{500}                                           & 368    \\
\hline

\end{tabular}
\medskip
\caption{\textbf{Comparison of TResNet to state-of-the-art models on transfer learning datasets (only ImageNet-based transfer learning results)}. Models inference speed is measured on a mixed precision V100 GPU. Since no official implementation of  Gpipe was provided, its inference speed is unknown.}
\label{Table:transfer_table}
\end{table}

%% file: multi_label.tex
For multi-label classification tests, we chose to work with MS-COCO dataset \cite{lin2014microsoft} (multi-label recognition task).
We used the 2014 split, which contains about 82K images for training and 41K for validation. In total, images are involved
with $80$ object labels, with an average of $2.9$ labels per image.

Our training scheme is similar to the one used for single-label training. The main difference is the loss function, which
is adapted for a multi-label setting - we implemented a variant of the well known focal-loss \cite{lin2017focal}, where two different gamma values are used for positive and negative sample. This enables to better tackle the highly imbalanced nature of a multi-label dataset.

Following the conventional settings \cite{chen2019multilabel,Wang2019MultiLabelCW}, we report the main performance evaluation metric, mean average precision (mAP), but in addition state
average per-class precision (CP), recall (CR), F1 (CF1) and
the average overall precision (OP), recall (OR), F1 (OF1).

In Table \ref{Table:multi_label_table}, we present the transfer learning results of TResNet model and compare it to the known state-of-the-art model.

\input{Tables/multi_label_table.tex}

We can see from Table \ref{Table:multi_label_table} that the TResNet-based solution significantly outperforms previous top solution for MS-COCO multi-label dataset, increasing the known SOTA by a large margin, from $83.7$ mAP to $86.4$ mAP. All additional evaluation metrics also show improvement.

%% file: Tables/multi_label_table.tex
\setlength{\tabcolsep}{4.0pt}
\begin{table}[hbt!]

\centering
\refstepcounter{table}
\begin{tabular}{|c|l|llllll|}
\hline
Backbone & \multicolumn{1}{c|}{mAP} & \multicolumn{1}{c}{CP} & \multicolumn{1}{c}{CR} & \multicolumn{1}{c}{CF1} & \multicolumn{1}{c}{OP} & \multicolumn{1}{c}{OR} & \multicolumn{1}{c|}{OF1}  \\ 
\hline
KSSNet\cite{Wang2019MultiLabelCW}~  & 83.7                     & 84.6                   & 73.2                   & 77.2                    & 87.8                   & 76.2                   & 81.5                      \\ 
\hline
TResNet-L & \textbf{86.4}            & \textbf{87.6}          & \textbf{76.0}          & \textbf{81.4}           & \textbf{88.4}          & \textbf{78.9}          & \textbf{83.4}             \\
\hline
\end{tabular}
\medskip
\caption{\textbf{Comparison of TResNet to state-of-the-art model on multi-label classification on MS-COCO dataset}. KSSNet \cite{Wang2019MultiLabelCW}, is the known SOTA, based on ResNet101 backbone.}
\label{Table:multi_label_table}
\end{table}

%% file: object_detection.tex
While our main focus was on various classification tasks, we wanted to further test TResNet on another popular computer vision task - object detection.

We used the known MS-COCO \cite{lin2014microsoft} dataset (object detection task), with a training set with 118k images, and an evaluation set (minival) of 5k images. For training, we used the popular mm-detection \cite{mmdetection} package, with FCOS \cite{tian2019fcos} as the object detection method and the enhancements discussed in ATSS \cite{zhang2019bridging}.

We trained with SGD optimizer for $70$ epochs with $0.9$ momentum, weight decay of $0.0001$ and batch size of $24$.  We used learning rate warm up, initial learning rate of 0.01 and 10x reduction at epochs 40, 60. We also implemented the data augmentations techniques described in \cite{Liu_2016}. 

For a fair comparison, we used first ResNet50 as backbone, and then replace it by TResNet-M (both give similar GPU throughput). Comparison results appear in Table \ref{Table:object_detection_table}.

\medskip
\input{Tables/object_detection_table.tex}
\medskip

We can see from Table \ref{Table:object_detection_table} that TResNet-M outperform ResNet50 on this object-detection task, increasing COCO mAP score from $42.8$ to $44.0$. This is consistent with the improvement we saw in the single-label ImageNet classification task.

%% file: Tables/object_detection_table.tex
\begin{table}[hbt!]
\centering
% \refstepcounter{table}
\begin{tabular}{|c|c|c|}
\hline
Method & Babkbone & mAP \%  \\ 
\hline
FCOS   & ResNet50 & 42.8    \\ 
\hline
FCOS   & TResNet-M & \textbf{44.0} \\  
\hline
\end{tabular}
\medskip
\caption{\textbf{Comparison of TResNet-M to ResNet50 on MS-COCO object detection task}. Results were obtained using  mm-detection   package,  with  FCOS  as the object detection method .}
\label{Table:object_detection_table}
\end{table}

%% file: Conclusion.tex
In this paper, we point out a possible blind-spot of latest developments in neural network design patterns. They tend not to consider actual GPU utilization as one of the measurements for a network quality. While GPU inference speed is sometimes measured, GPU training speed and maximal possible batch size are widely overlooked. 
For many real-world deep learning applications, training speed, inference speed and maximal batch size are all critical factors.

To address this issue, we propose a carefully selected set of design refinements, which are highly effective in utilizing typical GPU resources - SpaceToDepth stem cell, economical AA downsampling, Inplace-ABN operations, block-type selection redesign and optimized SE layers.
We combine these refinements with a series of code optimizations and enhancements to suggest a family of new models, dedicated for GPU high-performance, which we call TResNet. 

We demonstrate that on ImageNet, all along the top-1 accuracy curve TResNet gives better GPU throughput than existing models. In addition, on three commonly used downstream single-label classification datasets it reaches new state-of-the-art accuracies.
We also show that TResNet generalizes well to other computer vision tasks, reaching top scores on multi-label classification and object detection datasets.

%% file: appendix.tex
\appendix
\begin{appendices}
% \section{Overall Architecture of TResNet Models}
% \label{sec:Overall_archhitecture}
% \input{Tables/TResnet_layers_table}
\section{Code for Different Modules in TResNet}
\label{sec:Modouls_code}
\vspace{0.5cm}

\bf{JIT accelerated SpaceToDepth module}

\begin{python_l}
@torch.jit.script
class SpaceToDepthJIT(object):
    def __call__(self, x: torch.Tensor):
        N, C, H, W = x.size()
        x = x.view(N, C, H // 4, 4, W // 4, 4)  
        x = x.permute(0, 3, 5, 1, 2, 4).contiguous() 
        x = x.view(N, C * 16, H // 4, W // 4)
        return x
\end{python_l}

\vspace{0.5cm}
\bf{JIT accelerated AA downsampling module}
\begin{python_l}
@torch.jit.script
class AADownsamplingJIT(object):
    def __init__(self, channels: int, mixed_precision: bool = True):
        a = torch.tensor([1., 2., 1.])
        filt = (a[:, None] * a[None, :]).clone().detach()
        filt = filt / torch.sum(filt)
        self.filt = filt[None, None, :, :].repeat((channels, 1, 1, 1))
        self.filt=self.filt.cuda()
        if mixed_precision:
                self.filt = self.filt.half()

    def __call__(self, input: torch.Tensor):
        input_pad = F.pad(input, (1, 1, 1, 1), 'reflect')
        return F.conv2d(input_pad, self.filt, stride=2, 
                        padding=0, groups=input.shape[1])
\end{python_l}
\clearpage

% \vspace{0.5cm}
\bf{Fast implementation of global average pooling}
\begin{python_l}
class FastGlobalAvgPool2d():
    def __init__(self, flatten=False):
        self.flatten = flatten

    def __call__(self, x):
        if self.flatten:
            in_size = x.size()
            return x.view((in_size[0], in_size[1], -1)).mean(dim=2)
        else:
            return x.view(x.size(0), x.size(1), -1).mean(-1).view(
                          x.size(0), x.size(1), 1, 1)
\end{python_l}

\end{appendices}